\documentclass[letterpaper, 10 pt, conference]{ieeeconf/ieeeconf}  
\usepackage{graphicx}
\usepackage{afterpage}
\usepackage{amsmath}
\usepackage{amsfonts}
\usepackage{balance}

\usepackage[caption=false, font=footnotesize]{subfig}
\usepackage[font=small]{caption}
\usepackage[font=small]{subcaption}

\IEEEoverridecommandlockouts                              

\usepackage{booktabs}   
\usepackage{siunitx}    
\usepackage{algorithm}
\usepackage{algpseudocode}
\usepackage{url}
\usepackage{hyperref}
\usepackage{tikz}
\usetikzlibrary{positioning,arrows.meta}
\usepackage[most]{tcolorbox}
\tcbset{
  promptbox/.style={
    listing only,
    sharp corners,
    boxrule=0.4pt,
    colback=gray!5,
    colframe=black,
    left=1em, right=1em, top=0.8em, bottom=0.8em,
    fonttitle=\bfseries,
    before skip=1em, after skip=1em,
  }
}
\usepackage[caption=false,font=footnotesize]{subfig}

\newcommand{\BubbleWidth}{7.2cm}   
\newcommand{\BubbleHeight}{0.8cm}   
\newcommand{\IconTextSep}{6pt}      

\usetikzlibrary{positioning,arrows.meta}

\tikzset{
  bubble/.style={
    draw,
    rounded corners=6pt,
    line width=0.8pt,
    minimum width=\BubbleWidth,
    minimum height=\BubbleHeight,
    inner sep=6pt,
    align=left,
    fill=white, 
  },
  flowarrow/.style={-Latex, line width=0.8pt}
}

\newcommand{\iconbubble}[3][\IconHeight]{%
  \begin{minipage}[c]{\BubbleWidth}%
    \setlength{\fboxsep}{0pt}%
    \begin{minipage}[c]{#1}%
      \centering
      \includegraphics[height=#1]{#2}%
    \end{minipage}\hspace{\IconTextSep}%
    \begin{minipage}[c]{\dimexpr\BubbleWidth-#1-\IconTextSep-1em\relax}%
      #3%
    \end{minipage}%
  \end{minipage}%
}

\title{\LARGE \bf
Efficient Navigation in Unknown Indoor Environments with Vision-Language Models
}

\author{David Schwartz$^{1,2}$, Kota Kondo$^{2}$, and Jonathan P. How$^{2}$
\thanks{$^{1}$ETH Z\"{u}rich, Rämistrasse 101, 8092 Zürich, Switzerland. {\tt\small dschwartz@ethz.ch}}%
\thanks{$^{2}$Massachusetts Institute of Technology, Cambridge, MA 02139, USA. {\tt\small \{davids7, kkondo, jhow\}@mit.edu}}%
}

\begin{document}

\maketitle
\thispagestyle{empty}
\pagestyle{empty}

\begin{abstract}

We present a novel high-level planning framework that leverages vision-language models (VLMs) to improve autonomous navigation in unknown indoor environments with many dead ends. Traditional exploration methods often take inefficient routes due to limited global reasoning and reliance on local heuristics. In contrast, our approach enables a VLM to reason directly about occupancy maps in a zero-shot manner, selecting subgoals that are likely to yield more efficient paths. At each planning step, we convert a 3D occupancy grid into a partial 2D map of the environment, and generate candidate subgoals. Each subgoal is then evaluated and ranked against other candidates by the model. We integrate this planning scheme into DYNUS \cite{kondo2025dynus}, a state-of-the-art trajectory planner, and demonstrate improved navigation efficiency in simulation. The VLM infers structural patterns (e.g., rooms, corridors) from incomplete maps and balances the need to make progress toward a goal against the risk of entering unknown space. This reduces common greedy failures (e.g., detouring into small rooms) and achieves about 10\% shorter paths on average.
\end{abstract}


\section{Introduction}
\label{sec:introduction}

Autonomous mobile robots can be applied in numerous domains such as package delivery and search-and-rescue. 
In many of these applications, the robots need to operate in dense, cluttered, unknown environments where prior maps are not available.
When navigating such spaces, traditional approaches such as DYNUS~\cite{kondo2025dynus} often take inefficient routes by entering dead ends like small rooms, due to their lack of higher-level reasoning abilities.  
For instance, DYNUS simply projects the goal location onto a local map around the robot, and generates paths without global reasoning.

In recent years, many works have been proposed to improve the planning capabilities of robotic systems by employing large language models (LLMs) \cite{Tagliabue2023, latif20243pllmprobabilisticpathplanning, cui2023surveymultimodallargelanguage}, including in frontier exploration schemes \cite{yokoyama2023} and object-goal navigation \cite{Qu2024}. 
We aim to leverage the reasoning competencies of multi-modal LLMs, in particular VLMs, to achieve efficient routing in unknown indoor environments that typically contain many impasses. 
The novelty of our approach lies in allowing VLMs to directly reason about structural cues from incomplete maps, rather than relying on RGB images.
At each step, we convert the 3D occupancy grid into a voxelized 2D map, generate candidate subgoals, and ask the VLM to score directions in the context of the partial map (e.g., infer likely rooms/corridors from partial structure despite occlusions). 
We then aggregate multiple VLM queries, rank candidates by expected path efficiency, and select the top subgoal.

\section{Related Work}
\label{sec:related_work}

\subsection{Planning in Unknown Environments}

Traditional planners typically rely on local heuristics and frontier exploration~\cite{kondo2025dynus,ren2025safety,zhou2020ego,tordesillas2021faster,quan2025state,fan2025flying}, which often yield suboptimal paths in cluttered or confined unknown environments.
There exist several planning and exploration methods that leverage LLMs to give semantic guidance to the robot, such as LFG \cite{shah2023navigationlargelanguagemodels} and IPPON \cite{Qu2024}. Other approaches like VLMaps \cite{Huang2023} and Tag-Map \cite{Zhang2024} construct open-vocabulary semantic maps that are queryable by LLMs. Moreover, vision-language frontier maps (VLFM) \cite{yokoyama2023} builds a language-grounded value map directly from RGB images.
Other approaches instead aim at reconstructing unknown parts of the environment by using deep learning \cite{shrestha2019learned, schmid2022sc, ramakrishnan2020occupancy, ericson2024beyond, ho2024mapex}.
These methods usually assume very repetitive floor patterns, as well as high-quality occupancy grids (i.e. little noise in the LiDAR data). However, success in our use case is very binary: since our main concern is to avoid dead ends, plausible reconstructions that mistake whether a path leads to an impasse are not very useful. 

\subsection{Real-World Grounding of Foundation Models}

A key challenge in deploying foundation models (FMs) is enabling them to reason about the world through meaningful, structured representations of the environment. 
To compensate for the lack of real-world experience in models not specifically trained for robotics applications, various adaptation approaches have been developed. These include the use of grounding functions to decode the likelihood of whether a particular robot skill will succeed \cite{saycan2022arxiv} or to predict the next token from an open vocabulary \cite{huang2023groundeddecodingguidingtext}, the incorporation of grounded closed-loop feedback in robot planning \cite{Huang2023innermonologue}, and the generation of in-context prompts for high-level plans \cite{Song2023}. Moreover, other efforts aim at creating a better (open-vocabulary) spatial representation of the environment. This can be in the form of queryable maps \cite{Chen2023, Zhang2024}, value maps \cite{yokoyama2023, Qu2024}, VLMs \cite{Chen2024, gao2024physicallygroundedvisionlanguagemodels} or 3D Gaussian splatting maps \cite{jiang2024multimodalllmguidedexploration}. Kim et al. \cite{Kim2023} proposed to incorporate the contextual information of previous actions for planning, while the authors of \cite{Kim2024} presented an algorithm to update the map based on the robot's real-world experience. 

\subsection{Robot Navigation Using Foundation Models}

Several systems demonstrate the utility of foundation models for long-horizon navigation. Shah et al. \cite{Shah2021} developed a method that combines a learned policy with a topological graph to navigate towards a visually indicated goal. This work was extended to kilometer-scale navigation in ViKiNG \cite{Shah2022}. An adjacent approach, LM-Nav \cite{shah2022lmnav}, combines three large independently pre-trained models to enable long-horizon instruction following in the wild. ViNT \cite{shah2023vint} on the other hand employs a transformer-based architecture to perform visual navigation. NoMaD \cite{sridhar2023nomadgoalmaskeddiffusion}, makes use of a diffusion policy, which has the advantage of being able to represent complex and multi-modal distributions.

A different approach to robot navigation is video language planning \cite{du2023video}. The algorithm outputs a plan with detailed multimodal specifications that describe how to complete a task, which can then be translated into real robot actions via policies conditioned on frames of the generated video. Learning language-conditioned object navigation from videos is also the topic of LeLaN \cite{hirose24lelan} and Mobility VLA \cite{chiang2024mobilityvlamultimodalinstruction}. 
Moreover, videos do not only serve a purpose in training: Zhang et al. \cite{zhang2024navid} proposed NaVid, a video-based VLM for navigation from on-the-fly video streams from a monocular camera as input. 
Furthermore, it is also possible to train foundation models for navigation with data gathered in simulation \cite{Zeng2024}.

The outlined navigation foundation models provide end-to-end learning solutions to the planning problem, with a variety of applications. However, they exist as a replacement, not an add-on, to traditional trajectory planners, and do not explicitly tackle the problem of avoiding dead ends in partially known spaces.

\section{Methodology}
\label{sec:methodology}

\subsection{Problem Formulation}
Our goal is to enable autonomous robots to efficiently reach a specified target in initially unknown indoor environments, while avoiding dead ends that require backtracking. We formulate this as a goal-directed exploration problem under partial observability.
At each decision step, the high-level planner receives (1) the terminal goal position and (2) a 3D occupancy grid that is built online, and outputs a navigation subgoal that is likely to result in the most direct possible path to the goal.

\begin{figure}[h]
\centering
\begin{tikzpicture}[node distance=0.45cm]

\node[bubble] (s1) {\iconbubble{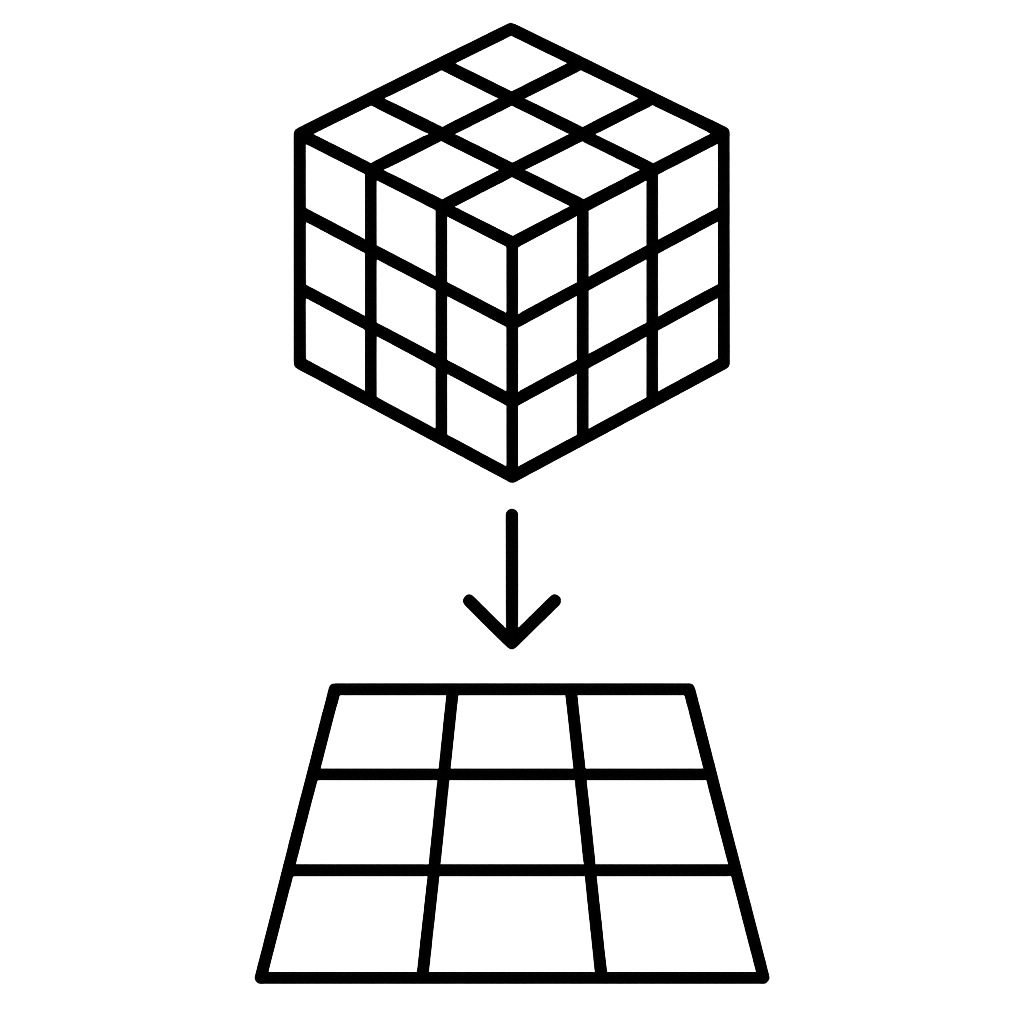}{
Convert 3D occupancy grid to 2D map
}};

\node[bubble, below=of s1] (s2) {\iconbubble{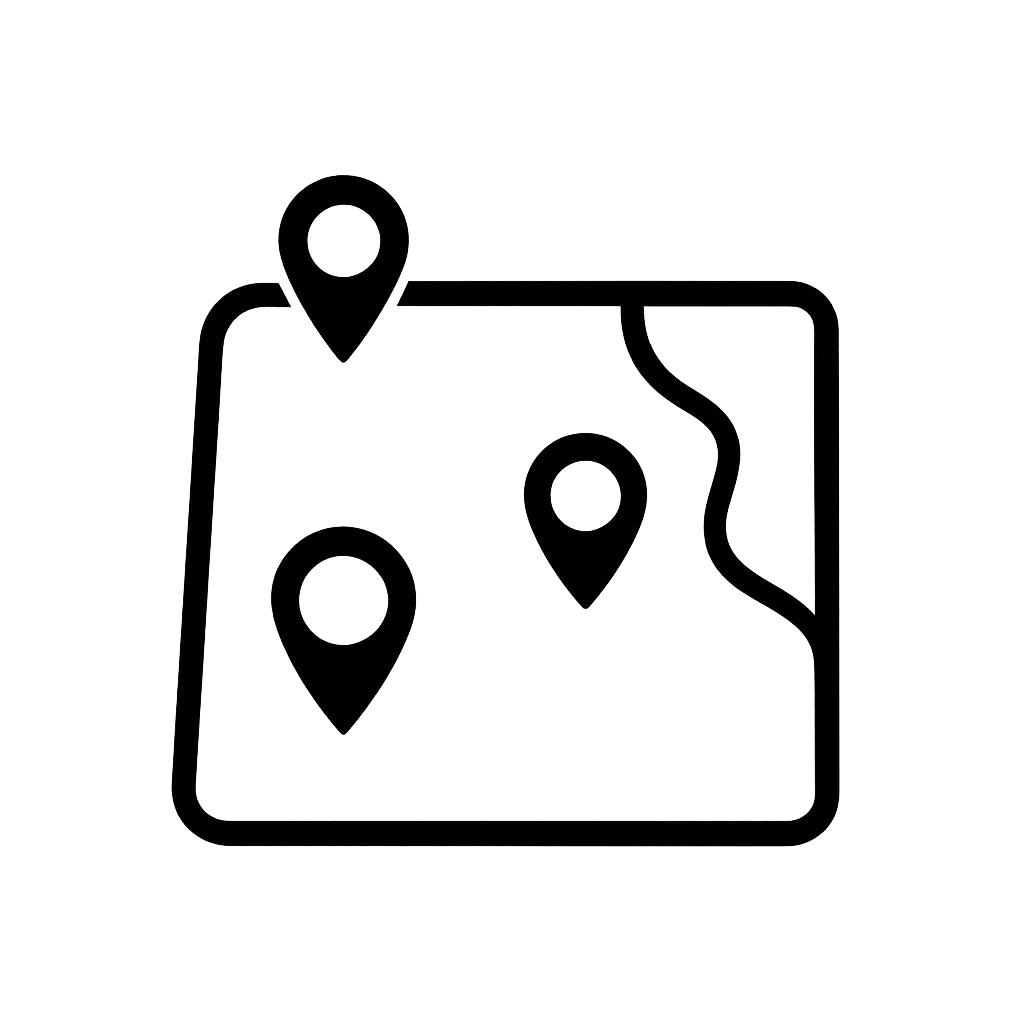}{
Generate candidate subgoals
}};

\node[bubble, below=of s2] (s3) {\iconbubble{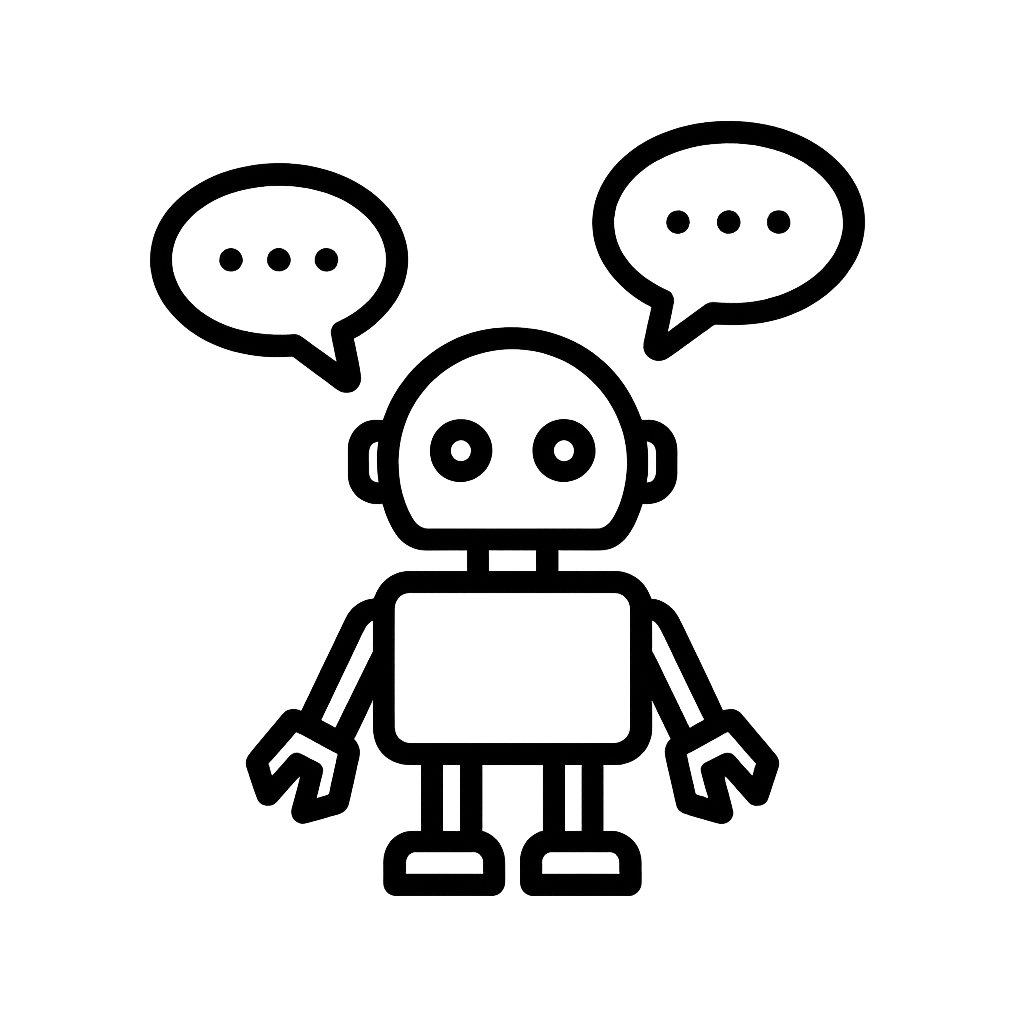}{
Sample \(n\) VLM responses with beliefs and justifications for each candidate, that choosing it is the best move
}};

\node[bubble, below=of s3] (s4) {\iconbubble{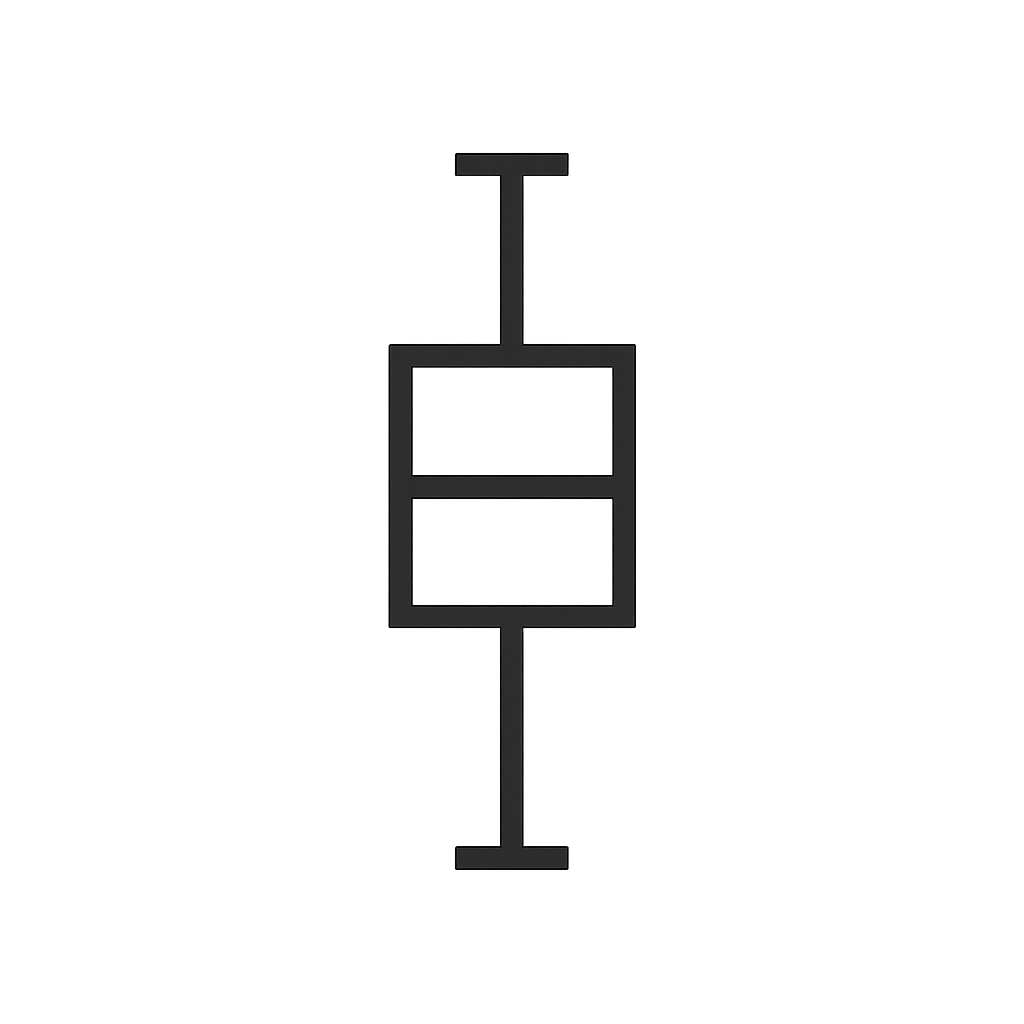}{
Compute normalized median belief and median absolute deviation for each of the candidates
}};

\node[bubble, below=of s4] (s5) {\iconbubble{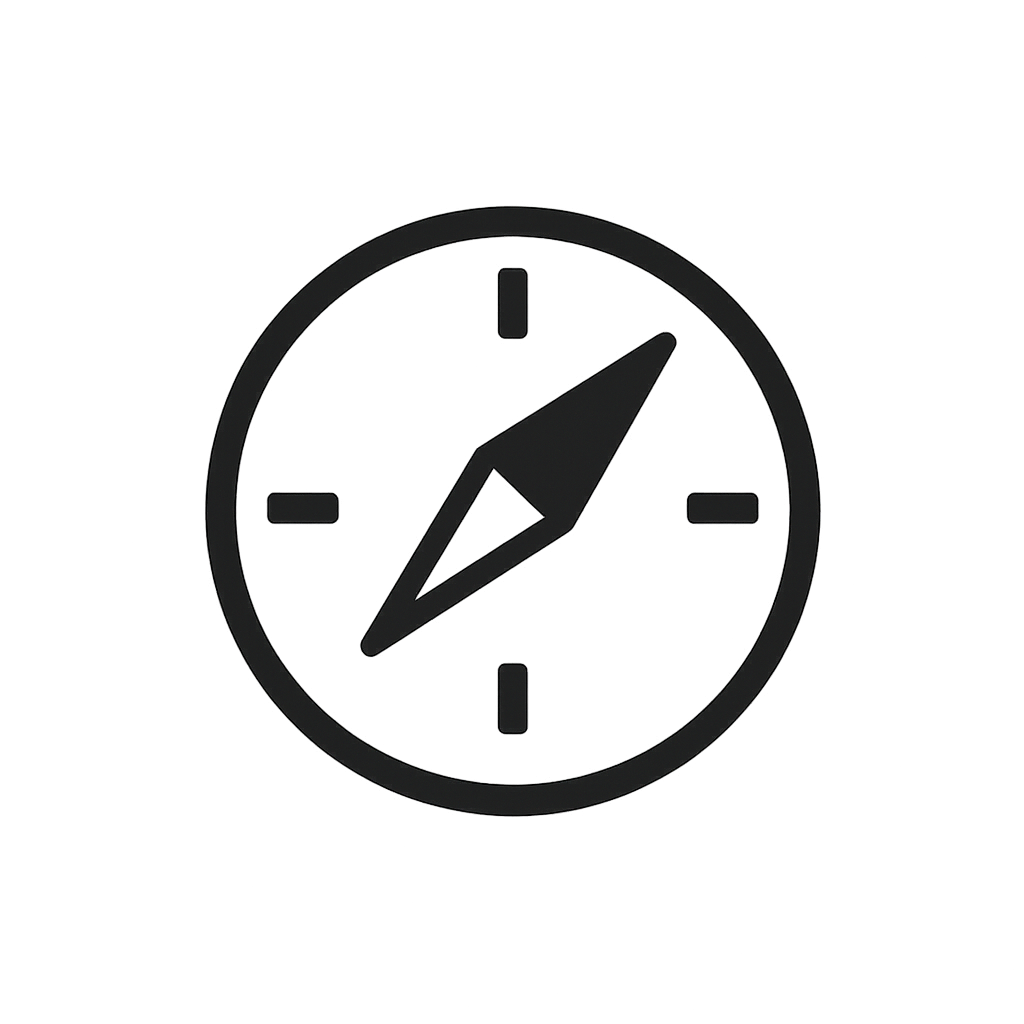}{
Choose a navigation subgoal
}};

\draw[flowarrow] (s1) -- (s2);
\draw[flowarrow] (s2) -- (s3);
\draw[flowarrow] (s3) -- (s4);
\draw[flowarrow] (s4) -- (s5);

\end{tikzpicture}

\caption{Overview of the main steps of the planning workflow.}
\end{figure}
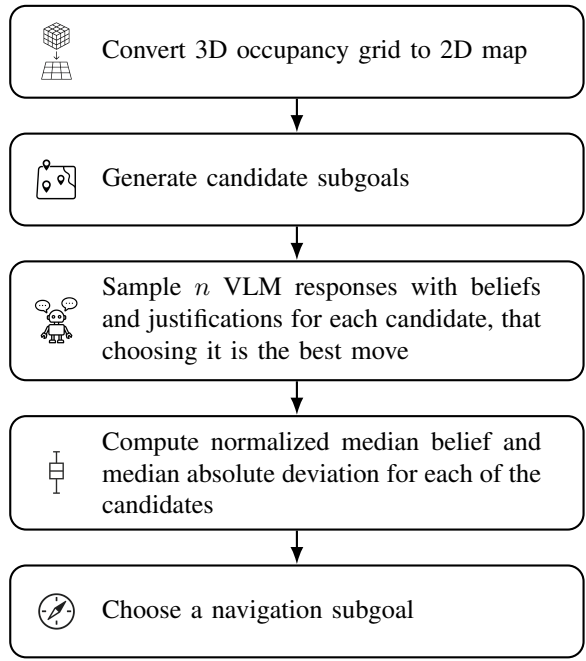


\subsection{Representing the Environment}

\begin{figure}[h]
    \centering
    \includegraphics[width=0.98\columnwidth]{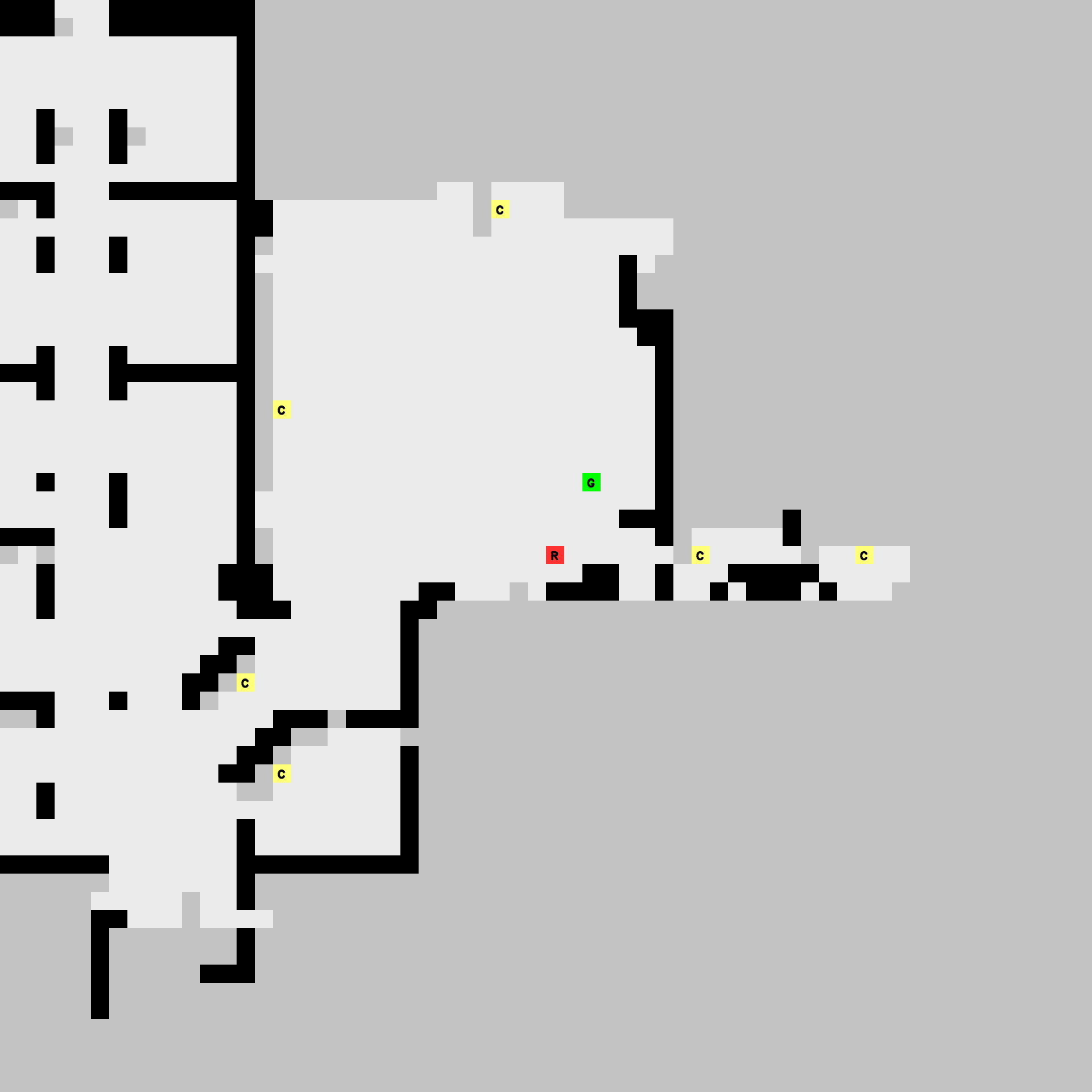}
    \caption{Image-based representation of the environment. The robot location is marked in red, the goal in green, candidate subgoals in yellow, free space in light gray, unknown space in dark gray, and occupied space in black.}
    \label{fig:example_img}
\end{figure}

The only information provided by the robot is a 3D occupancy grid of the environment. We do not rely on RGB images, which are constrained by lighting conditions and limited fields of view, but instead use occupancy data that capture the environment’s structure holistically and enable the system to operate effectively even in low-light conditions.
However, the vision-language model cannot parse and understand raw sensor data or a voxel map. 
To obtain useful navigation guidance from the VLM, we ground the VLM with a map-like representation: a projected 2D occupancy map annotated with the robot pose, the goal, and candidate subgoals. 
This makes spatial relations explicit so the VLM can reason about where to go next.
An example of such a representation is shown in Fig.~\ref{fig:example_img}.

\subsection{Converting a 3D Occupancy Grid to a 2D Map}

We keep a voxel map updated online.
The map encodes unknown/free/occupied cells and overlays the robot pose, the goal (or its projection), and candidate subgoals. A cell is marked as free only if there is sufficient $z$-axis clearance for the robot and is reachable. We pass the map to the model either as a string or as an image.
This can be achieved by associating each grid cell with a character or a color, respectively.
Because the input grid resolution is fixed, we specify the desired size of the map in both meters and grid cells, and down-sample the grid accordingly. To get the best possible approximation:

\begin{equation}
S_{\text{m}} \;=\; k \cdot \left( S_{\text{cells}} \cdot r \right), \quad k \in \mathbb{N},
\end{equation}
where $S_{\text{m}}$ is the desired map size in meters, $S_{\text{cells}}$ is the desired map size in number of grid cells, and $r$ is the occupancy grid resolution (meters per cell).


\subsection{Generating Candidate Subgoals}

We observed that it is easier for LLMs to evaluate candidate locations than it is to come up with subgoals on their own.
To generate subgoals (besides the terminal goal), we proceed as follows.
(i) \textbf{Frontiers:} mark as frontiers all free cells adjacent to at least one unknown cell (8-neighborhood).
(ii) \textbf{Clustering:} form 8-connected frontier clusters and pick, for each cluster, the cell nearest to the cluster centroid as a candidate subgoal.
(iii) \textbf{Deduplication:} remove near-duplicate or route-equivalent candidates {\textemdash} if two candidates are within a certain distance and lie on the same corridor/route, retain only one to avoid diluting VLM preferences with effectively identical options.
We also include the terminal goal (or its projection) as a candidate when it is reachable through free space.



\subsection{Instructing the Model}

We use the VLM as a scorer, not a policy. 
This lets the optimization-based trajectory planner, i.e. DYNUS~\cite{kondo2025dynus}, handle navigation in unknown space with safety guarantees, while the VLM provides high-level reasoning. 
Given candidate subgoals, we prompt the VLM to assign a relative likelihood to each candidate indicating how likely it is to lead to the goal faster than any of the other options.
This way, the subgoals are mutually exclusive and the assigned probabilities must sum to 1. We obtain a ranking of the candidates.
For interpretability, the VLM also returns a short rationale for each decision.

In addition to supplying a task description, we perform prompt engineering to help the model understand how to analyze the map, and how to make good planning decisions based on the analysis. The aim is to provide helpful guidelines without inducing any undesired behaviors, like being overly conservative around unknown parts of the map. 

\subsection{Selecting a Subgoal}

LLMs are stochastic and can hallucinate. 
To improve robustness and interpretability, we sample $n$ different outputs from our model at each planner query. We then consolidate the obtained answers by taking the median belief assigned to each of the candidates and normalizing it:

\begin{align}
    b_i &= \operatorname{median}\left( p_{i1}, p_{i2}, \ldots, p_{in} \right)\\
    \hat{b}_i &= \frac{b_i}{\sum\limits_{l=1}^n b_l}
\end{align}
The median has the advantage that it is robust to outliers, and does not blend different probabilities.
For instance, taking the mean of two valid but different solutions may not necessarily yield a good answer.
On top of that, we compute the median absolute deviation (MAD) for each belief as an uncertainty measure:

{\small \begin{equation}
    \mathrm{MAD}_i = \operatorname{median}\left( \left| p_{i1} - b_i \right|, \left| p_{i2} - b_i \right|, \ldots, \left| p_{in} - b_i \right| \right )
\end{equation}}

Finally, we select the candidate with the highest assigned belief as the next subgoal. 
In practice, we observed that while the relative likelihood assignments across different sample responses can vary, the orderings of the model's preferences rarely do. 
The main motivation behind the median filter is to use it as a safety mechanism that performs outlier rejection.

\section{Simulations}

\begin{figure*}[h]
    \centering
    \subfloat[Only DYNUS \cite{kondo2025dynus}]{
        \includegraphics[width=0.48\textwidth]{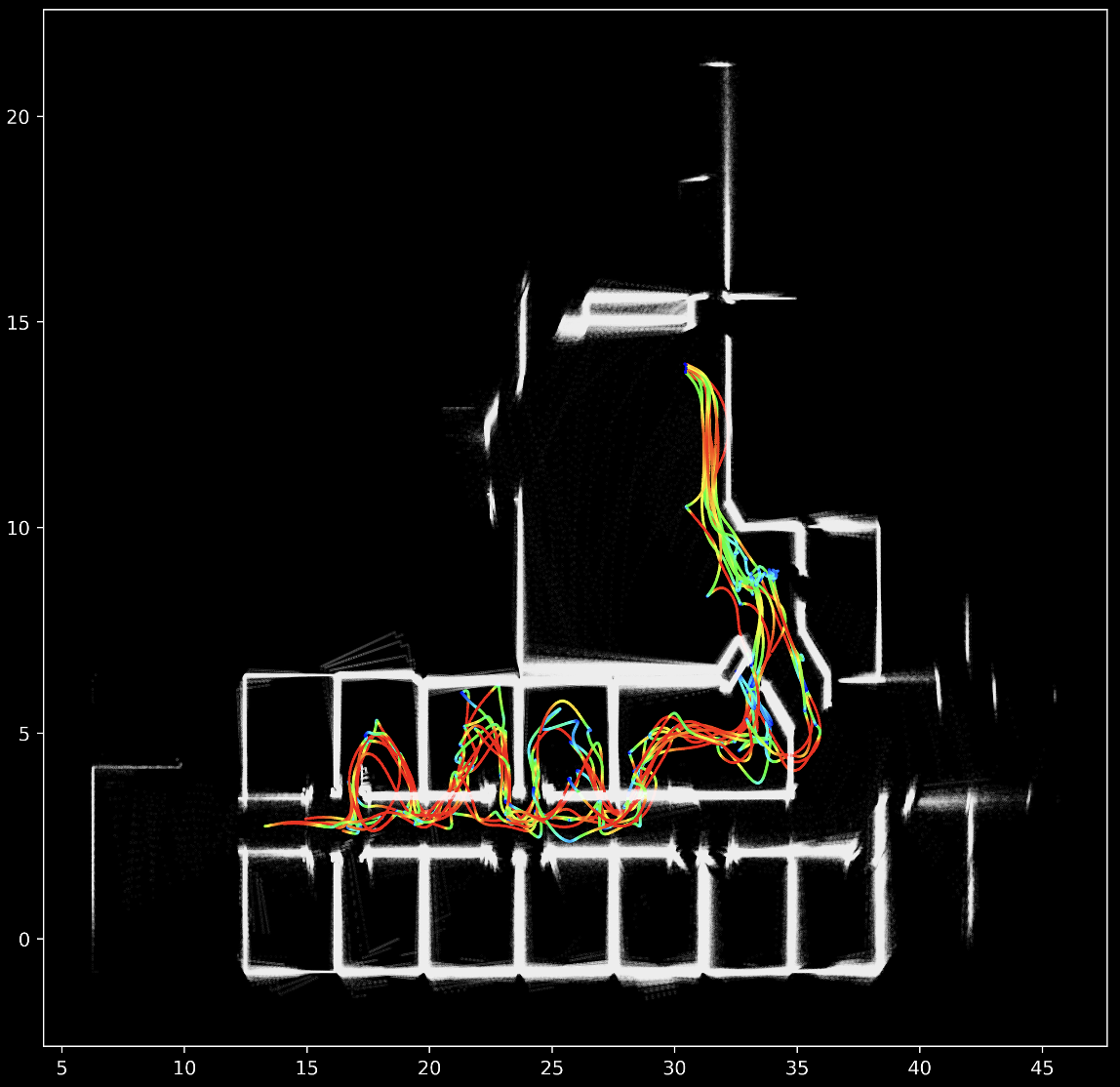}
        \label{fig:dynus_office}
    }
    \hfill
    \subfloat[DYNUS + GPT-4.1 (Ours)]{
        \includegraphics[width=0.48\textwidth]{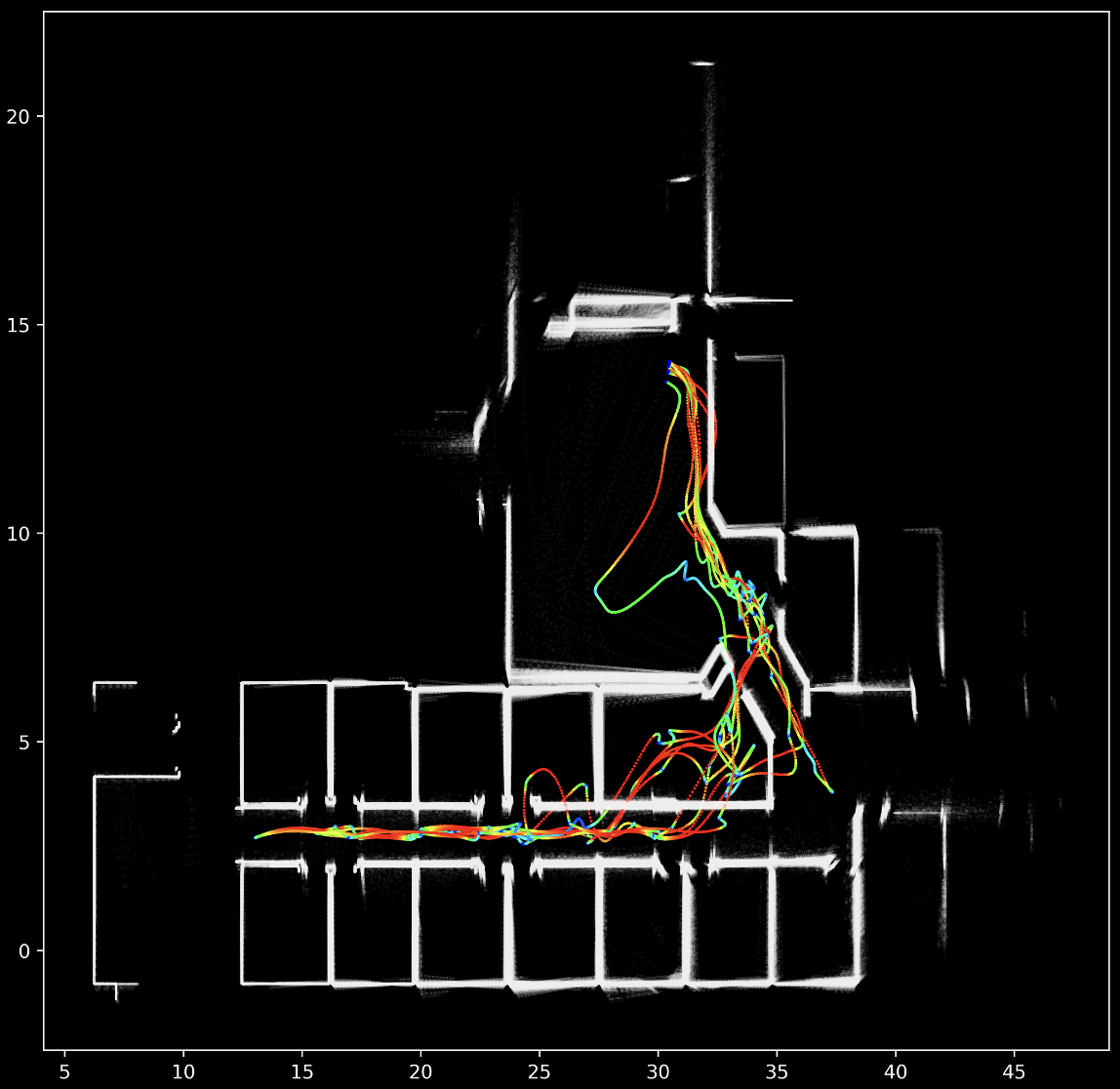}
        \label{fig:vlm_office}
    }
    \caption{Comparison of 15 paths produced by DYNUS~\cite{kondo2025dynus} with and without our proposed high-level planner in a Gazebo office environment. The start is at the bottom-left and the goal is at the top-right. Path color encodes speed (warmer colors indicate higher speed). The maximum velocity is set to 1.0 \SI{}{\meter\per\second}. Fig.~\ref{fig:dynus_office} shows that DYNUS alone often enters small rooms and backtracks when the goal is still far, leading to inefficient paths. Fig.~\ref{fig:vlm_office} shows that our method stays in hallways and avoids small rooms when far from the goal, yielding shorter paths.}
    \label{fig:comparison_dynus_office}
\end{figure*}

\subsection{Set-Up}

We compare our method against a baseline that uses only the traditional planner without our high-level module. 
We use the state-of-the-art trajectory planner DYNUS~\cite{kondo2025dynus} and run 15 trials in an office environment in Gazebo. 
Dynamic limits are set to 1.0 \SI{}{\meter\per\second}, 3.0 \SI{}{\meter\per\second\squared}, 4.0 \SI{}{\meter\per\second\cubed} for maximum speed, acceleration, and jerk, respectively.
We use the following configuration for the high-level planner:
\begin{itemize}
    \item model snapshot: \texttt{gpt-4.1-2025-04-14}\footnote{GPT-5 had not been released at the time this paper was written.},
    \item map format: image (performed better than strings),
    \item number of samples per query: \(n=3\),
    \item VLM input map size: \(60 \times 60\) cells covering \(24 \times 24\) meters.
\end{itemize}

\subsection{Results and Discussion}

Fig.~\ref{fig:comparison_dynus_office} shows the real-time point cloud produced by the mapper and the resulting paths. 
DYNUS alone often enters dead ends and backtracks.
In contrast, our planner avoids going into side rooms when the robot is far from the terminal goal. 
Note that some trajectories may appear inefficient; however, this is usually caused by poorly placed candidate subgoals rather than the scoring mechanism. 
Table~\ref{tab:results} reports the average path length: our method reduces path length by about 10\% relative to DYNUS. 

\begin{table}[h]
  \centering
  \resizebox{\columnwidth}{!}{
  \begin{tabular}{
        l
        c  
    }
    \toprule
    \textbf{Algorithm} 
      & \multicolumn{1}{c}{\textbf{Avg. Path Length [m]}} \\
    \midrule
    DYNUS \cite{kondo2025dynus} & 42.0 $\pm$ 3.2\\
    DYNUS + GPT-4.1 (Ours) & 37.9 $\pm$ 3.8\\
    \bottomrule
  \end{tabular}}
  \caption{Average path length (\(\pm\) standard deviation) from start to goal over 15 runs in a Gazebo office environment, comparing DYNUS~\cite{kondo2025dynus} with and without our high-level planner. Our method reduces path length by approximately 10\%.}

  \label{tab:results}
\end{table}

In addition to demonstrating the planning capabilities of multimodal LLMs, this experiment allowed us to study how they interpret partial maps for planning purposes. GPT-4.1's responses showed that it can identify structures such as rooms, hallways, and dead ends in floor plans, and balance exploration and exploitation by considering how much of the environment is unknown. 
The main bottleneck is the quality of the input map. 
When the map is dominated by unknown space or contains large errors, the VLM has little structure to reason about, which degrades performance.



\section{Conclusions}


This paper presents a VLM-based high-level planning framework that selects navigation subgoals at frontiers between known and unknown space in indoor environments. 
By providing occupancy maps as input, the model can reason about basic geometric structure (e.g., rooms and corridors). 
The method assumes minimal prior knowledge about the environment or sensors, and integrates cleanly with existing autonomy stacks.

In partially observed settings, reaching a goal while avoiding unnecessary detours or dead ends requires making decisions under uncertainty. 
Our simulations show that the proposed approach avoids most dead ends and yields shorter paths on average than the baseline in DYNUS~\cite{kondo2025dynus}. 
We also observe that VLMs can infer structural patterns from incomplete maps and make reasonable trade-offs when uncertainty is high.

The main limitation is model latency, which constrains real-time deployment. We expect this to improve with faster hardware and models. Future work could explore incorporating region-level semantic information to further inform decisions and support richer trade-offs.






\balance
\bibliographystyle{ieeeconf/IEEEtran}
\bibliography{references}

\begin{thebibliography}{10}
\providecommand{\url}[1]{#1}
\csname url@rmstyle\endcsname
\providecommand{\newblock}{\relax}
\providecommand{\bibinfo}[2]{#2}
\providecommand\BIBentrySTDinterwordspacing{\spaceskip=0pt\relax}
\providecommand\BIBentryALTinterwordstretchfactor{4}
\providecommand\BIBentryALTinterwordspacing{\spaceskip=\fontdimen2\font plus
\BIBentryALTinterwordstretchfactor\fontdimen3\font minus
  \fontdimen4\font\relax}
\providecommand\BIBforeignlanguage[2]{{%
\expandafter\ifx\csname l@#1\endcsname\relax
\typeout{** WARNING: IEEEtran.bst: No hyphenation pattern has been}%
\typeout{** loaded for the language `#1'. Using the pattern for}%
\typeout{** the default language instead.}%
\else
\language=\csname l@#1\endcsname
\fi
#2}}

\bibitem{kondo2025dynus}
\BIBentryALTinterwordspacing
K.~Kondo, M.~Peterson, N.~Rober, J.~R. Viso, L.~Jia, J.~Chen, H.~Merton, and
  J.~P. How, ``Dynus: Uncertainty-aware trajectory planner in dynamic unknown
  environments,'' 2025. [Online]. Available:
  \url{https://arxiv.org/abs/2504.16734}
\BIBentrySTDinterwordspacing

\bibitem{Tagliabue2023}
\BIBentryALTinterwordspacing
A.~Tagliabue, K.~Kondo, T.~Zhao, M.~Peterson, C.~T. Tewari, and J.~P. How,
  ``Real: Resilience and adaptation using large language models on autonomous
  aerial robots,'' 2023. [Online]. Available:
  \url{https://arxiv.org/abs/2311.01403}
\BIBentrySTDinterwordspacing

\bibitem{latif20243pllmprobabilisticpathplanning}
\BIBentryALTinterwordspacing
E.~Latif, ``3p-llm: Probabilistic path planning using large language model for
  autonomous robot navigation,'' 2024. [Online]. Available:
  \url{https://arxiv.org/abs/2403.18778}
\BIBentrySTDinterwordspacing

\bibitem{cui2023surveymultimodallargelanguage}
\BIBentryALTinterwordspacing
C.~Cui, Y.~Ma, X.~Cao, W.~Ye, Y.~Zhou, K.~Liang, J.~Chen, J.~Lu, Z.~Yang, K.-D.
  Liao, T.~Gao, E.~Li, K.~Tang, Z.~Cao, T.~Zhou, A.~Liu, X.~Yan, S.~Mei,
  J.~Cao, Z.~Wang, and C.~Zheng, ``A survey on multimodal large language models
  for autonomous driving,'' 2023. [Online]. Available:
  \url{https://arxiv.org/abs/2311.12320}
\BIBentrySTDinterwordspacing

\bibitem{yokoyama2023}
N.~Yokoyama, S.~Ha, D.~Batra, J.~Wang, and B.~Bucher, ``Vlfm: Vision-language
  frontier maps for zero-shot semantic navigation,'' in \emph{International
  Conference on Robotics and Automation (ICRA)}, 2024.

\bibitem{Qu2024}
\BIBentryALTinterwordspacing
K.~Qu, J.~Tan, T.~Zhang, F.~Xia, C.~Cadena, and M.~Hutter, ``Ippon: Common
  sense guided informative path planning for object goal navigation,'' 2024.
  [Online]. Available: \url{https://arxiv.org/abs/2410.19697}
\BIBentrySTDinterwordspacing

\bibitem{ren2025safety}
Y.~Ren, F.~Zhu, G.~Lu, Y.~Cai, L.~Yin, F.~Kong, J.~Lin, N.~Chen, and F.~Zhang,
  ``Safety-assured high-speed navigation for mavs,'' \emph{Science Robotics},
  vol.~10, no.~98, p. eado6187, 2025.

\bibitem{zhou2020ego}
X.~Zhou, Z.~Wang, H.~Ye, C.~Xu, and F.~Gao, ``Ego-planner: An esdf-free
  gradient-based local planner for quadrotors,'' \emph{IEEE Robotics and
  Automation Letters}, vol.~6, no.~2, pp. 478--485, 2020.

\bibitem{tordesillas2021faster}
J.~Tordesillas, B.~T. Lopez, M.~Everett, and J.~P. How, ``Faster: Fast and safe
  trajectory planner for navigation in unknown environments,'' \emph{IEEE
  Transactions on Robotics}, vol.~38, no.~2, pp. 922--938, 2021.

\bibitem{quan2025state}
F.~Quan, Y.~Shen, P.~Liu, X.~Lyu, and H.~Chen, ``A state-time space approach
  for local trajectory replanning of an mav in dynamic indoor environments,''
  \emph{IEEE Robotics and Automation Letters}, 2025.

\bibitem{fan2025flying}
X.~Fan, M.~Lu, B.~Xu, and P.~Lu, ``Flying in highly dynamic environments with
  end-to-end learning approach,'' \emph{IEEE Robotics and Automation Letters},
  2025.

\bibitem{shah2023navigationlargelanguagemodels}
\BIBentryALTinterwordspacing
D.~Shah, M.~Equi, B.~Osinski, F.~Xia, B.~Ichter, and S.~Levine, ``Navigation
  with large language models: Semantic guesswork as a heuristic for planning,''
  2023. [Online]. Available: \url{https://arxiv.org/abs/2310.10103}
\BIBentrySTDinterwordspacing

\bibitem{Huang2023}
C.~Huang, O.~Mees, A.~Zeng, and W.~Burgard, ``Visual language maps for robot
  navigation,'' in \emph{Proceedings - IEEE International Conference on
  Robotics and Automation}, vol. 2023-May, 2023.

\bibitem{Zhang2024}
M.~Zhang, K.~Qu, V.~Patil, C.~Cadena, and M.~Hutter, ``Tag map: A text-based
  map for spatial reasoning and navigation with large language models,''
  \emph{Conference on Robot Learning (CoRL)}, 2024.

\bibitem{shrestha2019learned}
R.~Shrestha, F.-P. Tian, W.~Feng, P.~Tan, and R.~Vaughan, ``Learned map
  prediction for enhanced mobile robot exploration,'' in \emph{2019
  International Conference on Robotics and Automation (ICRA)}.\hskip 1em plus
  0.5em minus 0.4em\relax IEEE, 2019, pp. 1197--1204.

\bibitem{schmid2022sc}
L.~Schmid, M.~N. Cheema, V.~Reijgwart, R.~Siegwart, F.~Tombari, and C.~Cadena,
  ``Sc-explorer: Incremental 3d scene completion for safe and efficient
  exploration mapping and planning,'' \emph{arXiv preprint arXiv:2208.08307},
  2022.

\bibitem{ramakrishnan2020occupancy}
S.~K. Ramakrishnan, Z.~Al-Halah, and K.~Grauman, ``Occupancy anticipation for
  efficient exploration and navigation,'' in \emph{Computer Vision--ECCV 2020:
  16th European Conference, Glasgow, UK, August 23--28, 2020, Proceedings, Part
  V 16}.\hskip 1em plus 0.5em minus 0.4em\relax Springer, 2020, pp. 400--418.

\bibitem{ericson2024beyond}
L.~Ericson and P.~Jensfelt, ``Beyond the frontier: Predicting unseen walls from
  occupancy grids by learning from floor plans,'' \emph{IEEE Robotics and
  Automation Letters}, 2024.

\bibitem{ho2024mapex}
C.~Ho, S.~Kim, B.~Moon, A.~Parandekar, N.~Harutyunyan, C.~Wang, K.~Sycara,
  G.~Best, and S.~Scherer, ``Mapex: Indoor structure exploration with
  probabilistic information gain from global map predictions,'' \emph{arXiv
  preprint arXiv:2409.15590}, 2024.

\bibitem{saycan2022arxiv}
M.~Ahn, A.~Brohan, N.~Brown, Y.~Chebotar, O.~Cortes, B.~David, C.~Finn, C.~Fu,
  K.~Gopalakrishnan, K.~Hausman, A.~Herzog, D.~Ho, J.~Hsu, J.~Ibarz, B.~Ichter,
  A.~Irpan, E.~Jang, R.~J. Ruano, K.~Jeffrey, S.~Jesmonth, N.~Joshi, R.~Julian,
  D.~Kalashnikov, Y.~Kuang, K.-H. Lee, S.~Levine, Y.~Lu, L.~Luu, C.~Parada,
  P.~Pastor, J.~Quiambao, K.~Rao, J.~Rettinghouse, D.~Reyes, P.~Sermanet,
  N.~Sievers, C.~Tan, A.~Toshev, V.~Vanhoucke, F.~Xia, T.~Xiao, P.~Xu, S.~Xu,
  M.~Yan, and A.~Zeng, ``Do as i can and not as i say: Grounding language in
  robotic affordances,'' in \emph{arXiv preprint arXiv:2204.01691}, 2022.

\bibitem{huang2023groundeddecodingguidingtext}
\BIBentryALTinterwordspacing
W.~Huang, F.~Xia, D.~Shah, D.~Driess, A.~Zeng, Y.~Lu, P.~Florence, I.~Mordatch,
  S.~Levine, K.~Hausman, and B.~Ichter, ``Grounded decoding: Guiding text
  generation with grounded models for embodied agents,'' 2023. [Online].
  Available: \url{https://arxiv.org/abs/2303.00855}
\BIBentrySTDinterwordspacing

\bibitem{Huang2023innermonologue}
W.~Huang, F.~Xia, T.~Xiao, H.~Chan, J.~Liang, P.~Florence, A.~Zeng, J.~Tompson,
  I.~Mordatch, Y.~Chebotar, P.~Sermanet, N.~Brown, T.~Jackson, L.~Luu,
  S.~Levine, K.~Hausman, and B.~Ichter, ``Inner monologue: Embodied reasoning
  through planning with language models,'' in \emph{Proceedings of Machine
  Learning Research}, vol. 205, 2023.

\bibitem{Song2023}
C.~H. Song, B.~M. Sadler, J.~Wu, W.~L. Chao, C.~Washington, and Y.~Su,
  ``Llm-planner: Few-shot grounded planning for embodied agents with large
  language models,'' in \emph{Proceedings of the IEEE International Conference
  on Computer Vision}, 2023.

\bibitem{Chen2023}
B.~Chen, F.~Xia, B.~Ichter, K.~Rao, K.~Gopalakrishnan, M.~S. Ryoo, A.~Stone,
  and D.~Kappler, ``Open-vocabulary queryable scene representations for real
  world planning,'' in \emph{Proceedings - IEEE International Conference on
  Robotics and Automation}, vol. 2023-May, 2023.

\bibitem{Chen2024}
\BIBentryALTinterwordspacing
B.~Chen, Z.~Xu, S.~Kirmani, B.~Ichter, D.~Driess, P.~Florence, D.~Sadigh,
  L.~Guibas, and F.~Xia, ``Spatialvlm: Endowing vision-language models with
  spatial reasoning capabilities,'' 2024. [Online]. Available:
  \url{https://arxiv.org/abs/2401.12168}
\BIBentrySTDinterwordspacing

\bibitem{gao2024physicallygroundedvisionlanguagemodels}
\BIBentryALTinterwordspacing
J.~Gao, B.~Sarkar, F.~Xia, T.~Xiao, J.~Wu, B.~Ichter, A.~Majumdar, and
  D.~Sadigh, ``Physically grounded vision-language models for robotic
  manipulation,'' 2024. [Online]. Available:
  \url{https://arxiv.org/abs/2309.02561}
\BIBentrySTDinterwordspacing

\bibitem{jiang2024multimodalllmguidedexploration}
\BIBentryALTinterwordspacing
W.~Jiang, B.~Lei, K.~Ashton, and K.~Daniilidis, ``Multimodal llm guided
  exploration and active mapping using fisher information,'' 2024. [Online].
  Available: \url{https://arxiv.org/abs/2410.17422}
\BIBentrySTDinterwordspacing

\bibitem{Kim2023}
B.~Kim, J.~Kim, Y.~Kim, C.~Min, and J.~Choi, ``Context-aware planning and
  environment-aware memory for instruction following embodied agents,'' in
  \emph{Proceedings of the IEEE International Conference on Computer Vision},
  2023.

\bibitem{Kim2024}
\BIBentryALTinterwordspacing
C.~Kim, K.~Kim, M.~Oh, H.~Baek, J.~Lee, D.~Jung, S.~Woo, Y.~Woo, J.~Tucker,
  R.~Firoozi, S.-W. Seo, M.~Schwager, and S.-W. Kim, ``E2map:
  Experience-and-emotion map for self-reflective robot navigation with language
  models,'' 2024. [Online]. Available: \url{https://arxiv.org/abs/2409.10027}
\BIBentrySTDinterwordspacing

\bibitem{Shah2021}
D.~Shah, B.~Eysenbach, G.~Kahn, N.~Rhinehart, and S.~Levine, ``Ving: Learning
  open-world navigation with visual goals,'' in \emph{Proceedings - IEEE
  International Conference on Robotics and Automation}, vol. 2021-May, 2021.

\bibitem{Shah2022}
D.~Shah and S.~Levine, ``Viking: Vision-based kilometer-scale navigation with
  geographic hints,'' in \emph{Robotics: Science and Systems}, 2022.

\bibitem{shah2022lmnav}
\BIBentryALTinterwordspacing
D.~Shah, B.~Osinski, B.~Ichter, and S.~Levine, ``{LM}-nav: Robotic navigation
  with large pre-trained models of language, vision, and action,'' in \emph{6th
  Annual Conference on Robot Learning}, 2022. [Online]. Available:
  \url{https://openreview.net/forum?id=UW5A3SweAH}
\BIBentrySTDinterwordspacing

\bibitem{shah2023vint}
\BIBentryALTinterwordspacing
D.~Shah, A.~Sridhar, N.~Dashora, K.~Stachowicz, K.~Black, N.~Hirose, and
  S.~Levine, ``Vi{NT}: A foundation model for visual navigation,'' in \emph{7th
  Annual Conference on Robot Learning}, 2023. [Online]. Available:
  \url{https://arxiv.org/abs/2306.14846}
\BIBentrySTDinterwordspacing

\bibitem{sridhar2023nomadgoalmaskeddiffusion}
\BIBentryALTinterwordspacing
A.~Sridhar, D.~Shah, C.~Glossop, and S.~Levine, ``Nomad: Goal masked diffusion
  policies for navigation and exploration,'' 2023. [Online]. Available:
  \url{https://arxiv.org/abs/2310.07896}
\BIBentrySTDinterwordspacing

\bibitem{du2023video}
Y.~Du, M.~Yang, P.~Florence, F.~Xia, A.~Wahid, B.~Ichter, P.~Sermanet, T.~Yu,
  P.~Abbeel, J.~B. Tenenbaum, \emph{et~al.}, ``Video language planning,''
  \emph{arXiv preprint arXiv:2310.10625}, 2023.

\bibitem{hirose24lelan}
N.~Hirose, C.~Glossop, A.~Sridhar, D.~Shah, O.~Mees, and S.~Levine, ``Lelan:
  Learning a language-conditioned navigation policy from in-the-wild video,''
  in \emph{Conference on Robot Learning}, 2024.

\bibitem{chiang2024mobilityvlamultimodalinstruction}
\BIBentryALTinterwordspacing
H.-T.~L. Chiang, Z.~Xu, Z.~Fu, M.~G. Jacob, T.~Zhang, T.-W.~E. Lee, W.~Yu,
  C.~Schenck, D.~Rendleman, D.~Shah, F.~Xia, J.~Hsu, J.~Hoech, P.~Florence,
  S.~Kirmani, S.~Singh, V.~Sindhwani, C.~Parada, C.~Finn, P.~Xu, S.~Levine, and
  J.~Tan, ``Mobility vla: Multimodal instruction navigation with long-context
  vlms and topological graphs,'' 2024. [Online]. Available:
  \url{https://arxiv.org/abs/2407.07775}
\BIBentrySTDinterwordspacing

\bibitem{zhang2024navid}
J.~Zhang, K.~Wang, R.~Xu, G.~Zhou, Y.~Hong, X.~Fang, Q.~Wu, Z.~Zhang, and
  H.~Wang, ``Navid: Video-based vlm plans the next step for vision-and-language
  navigation,'' \emph{Robotics: Science and Systems}, 2024.

\bibitem{Zeng2024}
K.-H. Zeng, Z.~Zhang, K.~Ehsani, R.~Hendrix, J.~Salvador, A.~Herrasti,
  R.~Girshick, A.~Kembhavi, and L.~Weihs, ``Poliformer: Scaling on-policy rl
  with transformers results in masterful navigators,'' 2024.

\end{thebibliography}


\clearpage
\nobalance



\renewcommand{\thefigure}{S\arabic{figure}}
\renewcommand{\thetable}{S\arabic{table}}
\renewcommand{\theequation}{S\arabic{equation}}
\setcounter{figure}{0}
\setcounter{table}{0}
\setcounter{equation}{0}

\section*{Supplementary Material}
\label{supplementary_material}

\subsection{Code}
\label{sec:code}

The implementation of our algorithm can be found at \href{https://github.com/mit-acl/dynus-vlm}{https://github.com/mit-acl/dynus-vlm}.

\subsection{\textsc{grid-world} Simulation Tool}
\label{sec:grid_world_tool}

\begin{figure*}[t]
    \centering
    \includegraphics[width=\linewidth]{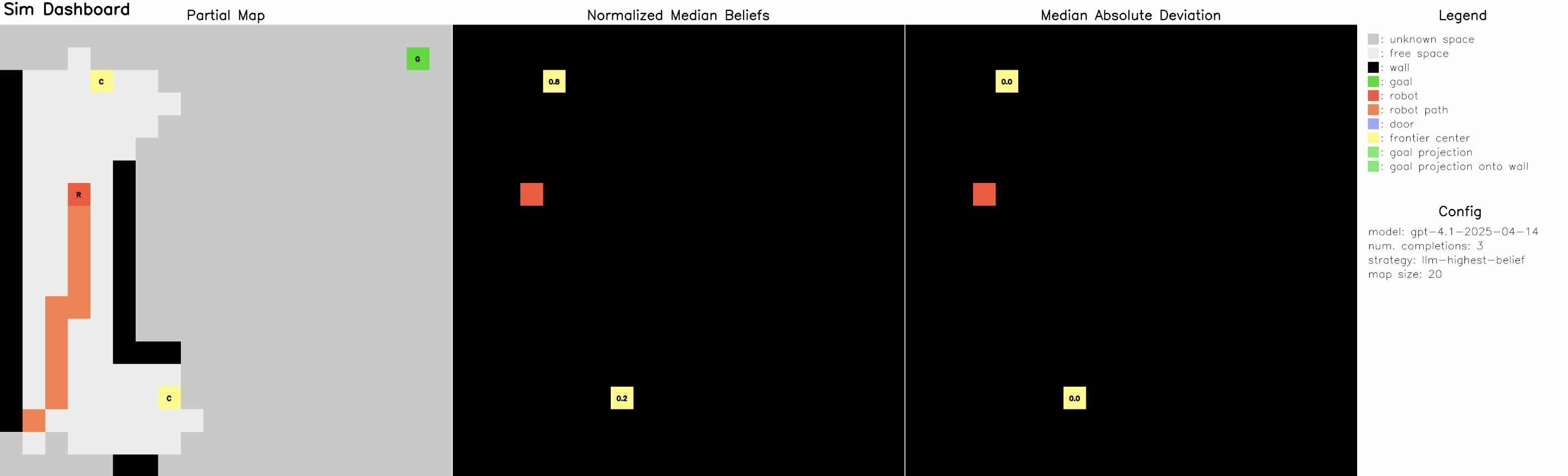}
    \caption{View of \textsc{grid-world}, a simulation tool we built to study the navigation capabilities of LLMs/VLMs. The 2D map on the left shows the robot (red) traversing a corridor in an effort to reach the final goal (green) in the top right. The squares in the middle and on the right display the normalized median belief and the median absolute deviation (MAD) of the belief, respectively, for each candidate subgoal the robot can move to. The implementation of \textsc{grid-world} is also included in the provided code.}
    \label{fig:grid_world}
\end{figure*}

See Fig.~\ref{fig:grid_world}.

\subsection{Prompt}
\label{sec:prompt}

\begin{tcolorbox}[title=Developer Prompt (1/2)]

\textbf{\large Identity}\\

\small{You are an expert in robotic navigation. Your job is to guide a mobile robot through an unfamiliar indoor environment, helping it reach a specified goal location using as few steps as possible.}\\
\\

\textbf{\large Task}\\

\small{For every step, given the context, you must:\\
- \textbf{Assign a belief value} (between 0 and 1) to each candidate move. This value reflects your confidence that this move is \textbf{strictly better}—meaning, more likely to reach the goal in fewer steps—compared to the other options.\\
- \textbf{Briefly justify} your belief value for each move.\\
\\}

\textbf{\large Workflow}\\

\textbf{\normalsize High-Level Problem-Solving Strategy}\\

\small{1. \textbf{Carefully evaluate each candidate move:}\\
   - Does it bring the robot closer to the goal without leading to a dead end (which would require backtracking)?\\
   - Is there any risk that this move leads to a dead end or gets trapped in a room?\\
   - Are there any visible or likely obstacles blocking progress in the direction of the goal?\\

2. \textbf{Compare all candidate moves:}\\
   - Explicitly compare the most promising candidates, focusing on enclosure risk, progress toward the goal, and likelihood of being blocked by walls or room boundaries.\\
   - If two candidates seem equally promising, prefer the one closer to the robot.\\

3. \textbf{Assign a belief value} to each candidate move, ensuring that the values \textbf{sum to exactly 1}.\\

---\\

\textbf{\normalsize 1. Candidate Move Evaluation}\\

\small{For each candidate move, do the following:\\

- \textbf{Explicit Wall Tracing:}\\  
\hspace*{1em} - Describe all the visible walls adjacent to each candidate and their immediate surroundings.

- \textbf{Path Feasibility Check:}\\
\hspace*{1em} - Is there a continuous open path from the candidate to the goal, or are there known walls that likely block it?

}
}
\end{tcolorbox}

\begin{tcolorbox}[title=Developer Prompt (2/2)]
\small{
- \textbf{Enclosure Check:}\\
\hspace*{1em} - Does the candidate appear to be in an enclosed area or room (i.e., surrounded on 3 or more sides by walls or boundaries)?\\

- \textbf{Corridor/Room/Door Classification:}\\
\hspace*{1em} - For each candidate, classify whether it is in a likely corridor, open space, or a room, and briefly explain your reasoning.

- \textbf{Progress Toward Goal:}\\
\hspace*{1em} - Does this move reduce the distance to the goal or its projection?

- \textbf{Uncertainty/Risk Reporting:}\\
\hspace*{1em} - Explicitly note if there is ambiguity in the environment or if the candidate may risk getting trapped.\\

---\\}

\textbf{\normalsize 2. Candidate Move Comparison}\\

\small{- \textbf{Explicitly compare} the most promising candidates in terms of enclosure risk, progress toward the goal, and likelihood of being blocked.\\
- If several candidates are similarly promising, prefer the one closer to the robot's current position.\\

---\\}

\textbf{\normalsize 3. Belief Assignment}\\

\small{- For each candidate, \textbf{assign a belief value} (between 0 and 1) reflecting your confidence, grounded in your reasoning above.\\
- All belief values \textbf{must sum to exactly 1}.\\
\\}

\textbf{\large Provided Context}\\

\small{You will be provided, at each step:\\

- A \textbf{partially revealed grid map} (see legend below).\\
- \textbf{Robot position}: (row, col)\\
- \textbf{Goal (or goal projection) position}: (row, col)\\
- \textbf{Candidate moves}: Dictionary of candidate next-step coordinates.\\

\textbf{Grid map legend}:\\
- \textbf{Gray}: Unknown (unexplored)\\
- \textbf{White}: Free space (navigable)\\
- \textbf{Black}: Wall (obstacle)\\
- \textbf{Green / "G"}: Goal\\
- \textbf{Red / "R"}: Robot\\
- \textbf{Yellow / "C"}: Frontier (boundary between explored/unexplored)\\
- \textbf{Light Green / "P"}: Goal projection (estimated direction)\\
- \textbf{Light Green / "K"}: Goal projection onto wall\\
\\

\textbf{Tip:}\\
If any aspect of the candidate's local geometry is ambiguous, or if there is uncertainty about potential passages, say so explicitly and adjust your confidence accordingly.}
\end{tcolorbox}
\emph{Remark}: we feed in the prompt in markdown format. It was edited here for better readability.




\balance
\subsection{Example Responses}
\label{sec:example_responses}

We found that GPT-4.1 can:

\begin{itemize}
    \item identify structures such as rooms and hallways in a floor plan,
        \begin{quote}\small
        ``This candidate is to the far north, up an extended corridor and within a semi-enclosed room. There is no visible path southward toward the projected goal. The route is circuitous and likely inefficient, with higher enclosure risk.''
        \end{quote}
        \begin{quote}\small
        ``Southernmost exit from central area, leading south and slightly east. Appears to open onto a possible corridor or hallway extending south, roughly matching the direction of the goal. No signs of imminent enclosure.''
        \end{quote}
    \item identify dead ends,
        \begin{quote}\small
        ``Northeast frontier, but accessing it requires backtracking through already-explored space. No clear indication it leads toward the goal. Likely a dead-end room given the local geometry and surrounding walls.''
        \end{quote}
        \begin{quote}\small
        ``This candidate is located at the top-left part of the explored area, apparently in a cul-de-sac (enclosed by walls on north and west sides). No visible path toward the goal projection to the far southeast. High risk of being a dead end.''
        \end{quote}
    \item balance exploration and exploitation by considering how much of the environment is unknown,
        \begin{quote}\small
        ``This candidate is north of the robot and to the east, leading into a corridor in a section with more open unexplored area. There's a good chance it opens up, but it's not the closest to the direction of the goal projection.''
        \end{quote}
        \begin{quote}\small
        ``This is the southernmost candidate, ending the current visible main corridor. It is the frontier closest to the projected goal position, opens directly onto unknown space, and is highly promising for discovering the route to the goal. However, possible enclosure just south, so there is still risk.''
        \end{quote}
    \item reason about what unknown space might contain.
        \begin{quote}\small
        ``This is to the south of the current room and could be a passageway, but still not fully aligned with the more southeast position of the goal. Low enclosure risk, but the area seems more like an open room than a corridor.''
        \end{quote}
        \begin{quote}\small
        ``This move goes south, away from the central room, and may be a corridor leading farther south. Though not directly leading southeast, it's the only exit on the lower half; if this is a corridor, it could bend toward the east, following the general direction of the goal projection. Medium uncertainty.''
        \end{quote}
\end{itemize}

\end{document}